# AN ANT COLONY OPTIMIZATION ALGORITHM FOR JOB SHOP SCHEDULING PROBLEM

Edson Flórez[1], Wilfredo Gómez[2] and MSc. Lola Bautista[3]

Universidad Industrial de Santander, Bucaramanga, Colombia

[1]Systems Engineering Student
edson.florez@correo.uis.edu.co
[2]Systems Engineer, member of the Research Group in Biomedical Engineering
wilfredo.gomez@correo.uis.edu.co
[3]Director of the Research Group in Biomedical Engineering
lxbautis@uis.edu.co

*ABSTRACT*

*The nature has inspired several metaheuristics, outstanding among these is Ant Colony Optimization (ACO), which have proved to be very effective and efficient in problems of high complexity (NP-hard) in combinatorial optimization. This paper describes the implementation of an ACO model algorithm known as Elitist Ant System (EAS), applied to a combinatorial optimization problem called Job Shop Scheduling Problem (JSSP). We propose a method that seeks to reduce delays designating the operation immediately available, but considering the operations that lack little to be available and have a greater amount of pheromone. The performance of the algorithm was evaluated for problems of JSSP reference, comparing the quality of the solutions obtained regarding the best known solution of the most effective methods. The solutions were of good quality and obtained with a remarkable efficiency by having to make a very low number of objective function evaluations.*

## 1. INTRODUCTION

ACO is a metaheuristic that brings together concepts from fields such as Artificial Intelligence and Biology, inspired in the collective behavior of ants. These social insects form colonies of ants, which are self-organizing systems and decentralized which are considered as a Swarm Intelligence [12]. Thanks to that intelligence emerging from simple relationships between ants, a colony can solve complex problems in their environment, such as the problem of finding the shortest path between the colony and the food, which can be used to find the best solution for combinatorial optimization problems.

In this paper, we apply the collective intelligence of many simple agents to the problem of Job Shop Scheduling [22], which consists of finding an optimal plan that minimizes the makespan, which is the time required to perform a finite number of tasks in a finite number of machines [13]. Each task is a sequence of operations, each one with a determined machine and processing time. Feasible solutions must comply with the restrictions that apply to the problem of Job Shop Scheduling, as respecting the precedence between operations determining the technological sequence without interrupting any operations until completion [21]. The operations conform the graph nodes that represent the problem, united by edges in which ants are moving. Each individual only has local information of the system that shares through a hormone called pheromone.

The update of the pheromone trail deposited on the edges can be done globally or locally. Ants build roads that represent feasible solutions, guided by the pheromone trails and the heuristic information of each edge [1]. For this reason the ant population performs a stochastic search, selecting the next node to visit only based on information available locally, used on a probabilistic approach where initially the ant decisions are completely random in the absence of pheromone trails.

In the literature, several algorithms have been proposed following the ACO probabilistic technique for finding approximate solutions to complex optimization problems. The first ACO algorithm was Ant System (AS), proposed by Marco Dorigo in 1991 [26], and completed with the contributions of Maniezzo and Colorni [1]. New developments gave better results, like Ant Colony System (ACS) [2], the Max-Min Ant System (MMAS) [7], the Rank-based Ant System ($AS_{rank}$) [8], among others. This article presents a variant of Elitist Ant System, also proposed by Dorigo as an improvement to SH [1], applied in JSSP instances widely used known as LA instances, that were raised by Lawrence [11].

## 2. JSSP Problem Formulation

The JSSP or resource planning problem (or jobs) consists in "accommodate resources over time to perform a set of jobs" [6], building plan or execution sequence of jobs j in a set of m machines [13], where an operation is every job that is processed in each machine (Operation(j, m)) and is assigned a specific processing time.

This problem is presented in multiple human activities, taking applications to tasks such as scheduling for packet delivery (eg airway), computer networks (networking), computers (multitasking and multiprocessing), project management (agenda or plan), production and administrative processes (eg assembly lines, etc.) [20].

JSSP must comply with certain restrictions in the execution of jobs and the goal is to complete them in the shortest possible time. This time to optimize is known as makespan ($C_{MAX}$) or Maximum Workflow which forms the objective function to minimize given as $C_{MAX} = max_{i=1\ldots n}(C_i)$, where $C_i$ is the job $J_i$ completion time. It is a combinatorial optimization problem because the number of candidate solutions is combinatorial in size with variables of discrete nature, therefore the representation of the solutions are permutations over the operations of each job, making it impossible to determine all possible solutions in a reasonable time.

### 2.1. Computational complexity of JSSP

In 1976 Michael Garey [17] provided evidence that this problem is NP-hard for m> 2, ie cannot be quickly found (polynomial-time) an optimal solution for JSSP with more than two machines. Along with David Johnson in 1979, they finished demonstrating that JSSP is NP-hard [18], unless in Computational Complexity Theory is proved that P = NP, if so, any problem that can be checked quickly by a computer, it could also be quickly resolved by that computer.

The NP-hard complexity of JSSP lies in the vast number of possible combinations that arise because each sequence of operations on a machine can be permuted independently of the sequence of operations on another machine, so with a few jobs and machines can have $(j!)^m$ possible solutions which corresponds to the search space (S) of the problem.

### 2.2. Formal definition of Job Shop Scheduling Problem

Having [5]:

$J = \{j_1, j_2, j_3, \ldots, j_n\}$ : Set of *n* jobs to be processed.

$M = \{m_1, m_2, m_3, \ldots, m_m\}$ : Set of *m* machines or resources.

$O_{ik} = \{o_{i1}, o_{i2}, o_{i3}, \ldots, o_{ik_i}\}$ : Operation of the job $J_i$ that must be processed in the machine $M_k$ by $\tau_{ik}$.

$\tau_{ik} = \{\tau_{i1}, \tau_{i2}, \tau_{i3}, \ldots, \tau_{ik_i}\}$ : Uninterrupted period of processing time for each operation.

Objective function: Minimize $C_{MAX} = max(t_{ik} + \tau_{ik})$, $\forall \ (J_i \in J \ and \ M_k \in M)$

Subject to:

    Start times restriction for each operation $t_{ik} \geq 0$

    Precedence constraint $t_{ik} - t_{ih} \geq \tau_{ih}$ if $O_{ih}$ preceding $O_{ik}$

    Disjunctive restriction $t_{pk} - t_{ik} + K(1 - y_{ipk}) \geq \tau_{ik}$    $y_{ipk} = 1$ if $O_{ik}$ preceding $O_{pk}$,

                                 $t_{ik} - t_{pk} + K(y_{ipk}) \geq \tau_{pk}$    $y_{ipk} = 0$ in another case.

Where: $\{i, p\} \in J$ and $\{k, h\} \in M$, with $K > \sum_{i=1}^{n}(\sum_{k=1}^{m} \tau_{ik} - min(\tau_{ik}))$.

The previous set of constraints of the JSSP is explained of this way [21]:

Start restriction: The time when an operation starts are not specified, so work can start at any point in time as long as the required machine is available.

Restriction of precedence: Each job must go through a particular sequence of operations that is predefined, so that operations cannot begin until the end of its predecessor, preventing the processing of two operations of the same job simultaneously.

Restrictions disjunctive: A machine can process only one job at a time. Each operation must be fully processed on a single machine and cannot be interrupted even if there are jobs waiting for that machine to be available, for instance, no work may be processed more than once on the same machine.

In addition to the above restrictions, we have determined that all operations have the same priority of processing, and all machines are the same and can be idle at any time. The fulfillment of these restrictions can be seen clearly by a Gantt chart (Figure 1), which shows an instance of JSSP (Table 1) matrix defined by [15], in which it has an additional column to indicate that each row of the matrix corresponds to a job (J1, J2 and J3).

Table 1. An instance of JSSP 3×3

| Job (J) | Machine (*time*) | | |
|---|---|---|---|
| Sequence: | S1 | S2 | S3 |
| J1 | 3 (4) | 2 (3) | 1 (3) |
| J2 | 2 (1) | 3 (2) | 1 (4) |
| J3 | 2 (3) | 1 (2) | 3 (3) |

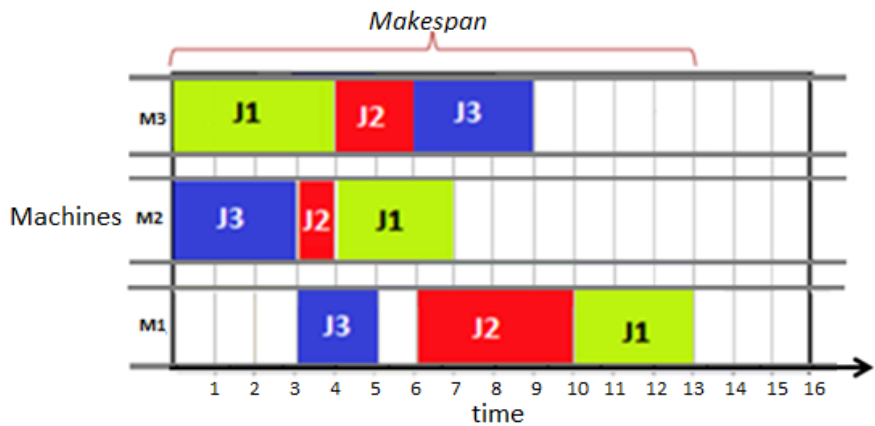

Figure 1. Gantt diagram of a 3×3 instance of JSSP

The JSSP is usually represented as a disjunctive graph $G = (V, C \cup D)$ [14], where $V$ is the set of nodes (Figure 2) representing the *Operations* (*job*, *machine*) with the exception of starting node (I) and ending nodes (F) of the graph, $C$ is a set of directed graphs (→) linking operations corresponding to the same job (technological sequence), and $D$ is a set of undirected graphs (←---→) connecting operations running on a same machine. In addition the processing time of each operation is placed in the upper part of node.

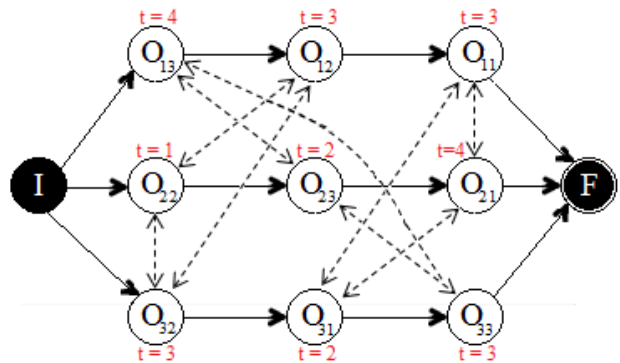

Figure 2. Graph of a 3×3 instance of JSSP

The problem of Job Shop Scheduling has been tackled with methods that can only solve instances of a limited number of operations, because they perform exhaustive searches to find the exact solution, as Branch and Bound (B&B) proposed in 1960 [23], can solve only up to 15 x 14, ie up to 220 operations [24]. So it must use approximate methods (Table 2 [9]) like simulated annealing (SA), Tabu Search (TS) [10], Iterative Local Search (ILS), GRASP, ACO, Evolutionary Algorithms (EA) as Artificial Immune System (AIS) and Cultural algorithm (CULT), etc.

Table 2. The main features of metaheuristics

| Metaheuristic | Features |
|---|---|
| SA | Acceptance criteria <br> Cooling Time |
| TS | Choosing neighbor (tabu list) <br> Suction Criterion |
| EC | Recombination <br> Mutation <br> Selection |
| ILS | Local search <br> Initial movement <br> Acceptance criteria |
| ACO | Construction probabilistic <br> Update pheromone |
| GRASP | Local Search <br> Restricted Candidate List (RCL) |

## 3. ANT COLONY OPTIMIZATION

This bioinspired algorithm is based on a population of ants that perform a cooperative search. In an experiment of the self-organization of Argentine ants made in 1989 [12], we observed the feeding behavior of a colony of ants, that were able to find the shorter branches of a bridge between the nest and the food (Figure 3), through the pheromone trail they leave behind when moving.

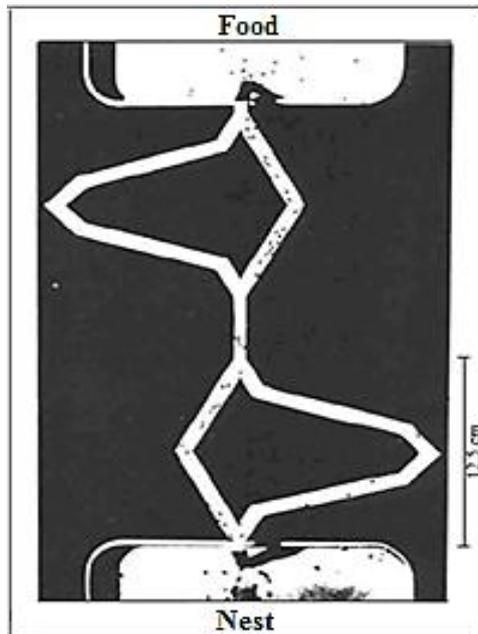

Figure 3. Picture of a colony of ants that find the shortest path to the food [12]

The ants initially move randomly in search of food and along the way back to the colony the pheromone is deposited. If another ant finds this trail, probably it will follow it increasing the amount of pheromone, which further stimulates other ants to follow this path (Figure 4). But over time the pheromone trail starts to evaporate and reduces its attractiveness, making more attractive only the most used trajectories, causing convergence to an optimal solution that is the only path that eventually most ants will follow. By the long road less pheromone accumulates because of the low passing frequency of the ants when they spend more time completing their road.

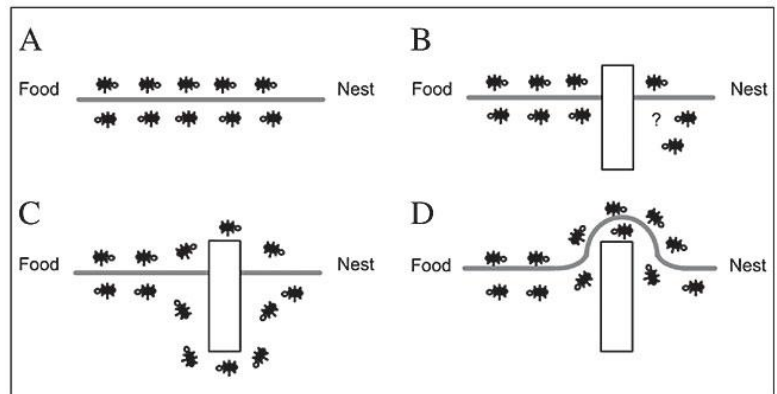

Figure 4. A. ants in a pheromone trail between nest and food; B. an obstacle interrupts the trail; C. ants find two paths to go around the obstacle; D. a new pheromone trail is formed along the shorter path [19]

In the ACO algorithms family, ant's behavior is simulated with a virtual agent that has the capacity to explore a limited search space and obtain information about the surrounding environment. The artificial ant (k) moves from one node to another (from source node i to destination node j), building step by step solution to be written to the $Tabu_k$ memory (that stores information about the nodes sequence or route taken until time t), that ends when it reaches one of the accepting states defined by the objective of the problem.

Thus, the ants can construct approximate solutions to complex problems such as sequencing, assigning, planning or programming. Each edge of the graph has two types of associated information that guide the movement of the ant [4] and whose values are modified by ants at each iteration:

$\eta_{ij}$ Heuristic information that measures the heuristics preference of moving from node *i* to node *j*, when touring the edge $a_{ij}$. Ants do not change this information during the execution of the algorithm.

$\tau_{ij}$ Information of artificial pheromone trails, that measures the "desirability learned" of the *i* to *j* movement. This information is modified during the execution of the algorithm depending on the solutions found by the ants to reflect the experience gained by these agents.

Pseudocode of the ACO metaheuristic [3]:
*ACO procedure*
   Set parameters,Initialize the pheromone trails
       scheduled activities
           Construction of solutions by ants
          Server of actions (Optional)
         Updating pheromone
       End-Scheduled activities
*End-procedure*

The metaheuristic consists of a parameter initialization step and three algorithmic procedures whose activation is regulated by the builder *Scheduled activities*, in which is repeated until a termination condition is met, such as reaching a maximum number of iterations or a maximum CPU time. The three algorithmic procedures submitted to the *Scheduled activities* consist of [25]:

*Construction of solutions by ants* is the probabilistic construction of solutions by all the ants in a colony, which visit the adjacent states of the considered problem. The ants can move by applying a stochastic decision policy using information from the pheromone trails and the heuristic information, with which ants incrementally construct a solution to the problem.

*Server of actions* are centralized actions that modify the behavior of the algorithm and cannot be developed by ants individually. The most common is the local optimization or improvement of the solutions with the application of a local search algorithm. The locally optimized solutions are then used to set the values of the pheromone to update.

*Updating pheromone* is the process that updates the pheromone trails on each $a_{ij}$ edge, called *posteriori online update* or *offline* because it is performed at the end of a road. The amount of pheromone that deposits each ant at the edges depends on the total length of the path (equation 3). It also can perform a *step by step online update of the pheromone trails*, that is a local update or in "real time" of the pheromone, performed when an ant moves from node *i* to node *j*. The pheromone trail value is reduced by a constant evaporation of pheromone, which prevents premature convergence of the algorithm by discarding the less frequented corners.

## 4. ELITIST ANT SYSTEM (EAS)

This version of the ACO implements a simple change to the Ant System that improves the results, simply reinforcing the pheromone trail of the best path that is found in each iteration. At the edges of the best generated solution by an ant, more pheromone is deposited through all the other ants.

In this algorithm artificial ants perform a probabilistic construction of solutions in each cycle, for which they require represent the problem by means of a graph in which the ants move along each edge from one node to another to build roads that represent solutions from a randomly chosen initial node, the following choice is the next node in this path is done according to the state transition rule (equation 1).

$$p_{ij} = \frac{(\tau_{ij})^\alpha (\eta_{ij})^\beta}{\sum_{l \in Tabu_k} (\tau_{il})^\alpha (\eta_{il})^\beta}$$

$$si\ j, l \notin Tabu_k$$

**Equation 1**

Where α and β parameters determine the influence of the values of the pheromone information ($\tau$) and from the heuristic information (η) respectively, over the decision of each ant (k). It seeks that the edges with large amount of pheromone to be the most visible, having a higher transition probability to the edges of the other nodes of the set of achievable operations. To have a balanced algorithm (with an appropriate adjustment), α and β parameters must have appropriate values, avoiding close to zero values, because if α = 0, only the heuristic information would indicate that possible elements of the solution will have a higher probability of being selected, which corresponds to a stochastic greedy algorithm (greedy), and if β = 0, will only be relevant the amount of pheromone. In both cases the ants might get stuck in a local optimum, generating the same solution in each iteration, without opportunities to find a better solution which could be the global optimum solution. These parameters are normally set to integer values between 1 and 5, but in this case we will relate them as follows $\beta = (1 - \alpha)$ with $\alpha \in (0,1]$.

The amount of $\tau_{ij}(t)$ pheromone present at each edge of the road in the $t$ generation is given by the equation 2.

$$\tau_{ij}(t) = \sum_{k=1}^{n} \Delta \tau_{ij}^k + \rho * \tau_{ij}(t-1)$$

**Equation 2**

Where $\tau_{ij}^k(t)$ is the contribution of the $k$ ant to the total pheromone of the $t$ generation and $\rho$ is the evaporation rate of the pheromone. The reason for including the evaporation rate is that old pheromone should not have much influence on future decisions of the ants. The amount of pheromone that each ant is contributing depends on the quality of the solution obtained which is inversely proportional to the cost of the solution of the objective function (equation 3).

$$\Delta \tau_{ij}^k = \frac{Q}{L_k}$$

**Equation 3**

Where $Q$ is a constant and $L_k$ is the length of the makespan of the solution obtained by the $k$ ant.

To accelerate the convergence of the algorithm, increasing the visibility of the pheromone trail on all edges of the shortest path, passing all elitists ants (e) of the system. Therefore, the equation 3 for the best path built in each cycle is replaced by the equation 2.

$$\Delta \tau_{ij}^k = \frac{Q}{L_k} * e$$

**Equation 4**

## 5. EAS IMPLEMENTATION FOR JSSP

The rapid convergence of this algorithm can reduce the scanning capability since the ants soon will end in a single way, which can be a local optimum. To compensate this, is allowed to include in the set of achievable operations (point 3.3 of pseudocode), operations that makes the machines wait (on pause) some units of time to begin execution because the corresponding job is still active on another machine. But this operation that delay or retards the onset of the machines will only be selected if the edge that reaches the node, has enough pheromone to make the probability to be greater than the operations that have immediately available jobs. That will only be given with large amounts of pheromone, because having idle machines is not adequate and is penalized lowering the visibility of the operation.

This method further explores the search space in order to obtain many solutions, from which it can be obtained solutions that exceed the local optima found in the first iterations. These optimal are the ones limiting the search, stopping it on solutions distant up to a 5% the global optimum. The initial diversity of the algorithm is the one that ensures that the ants move towards the search space where the path corresponding to the overall optimal solution is found. The following is the pseudocode implemented to solve the JSSP:

## EAS procedure

1. *Parameter initialization:* $\alpha, \beta, \rho, K$ *Ants and C Cycles*
2. *For every* $a_{ij}$ *edge, do:*
   $\tau_{0_{ij}} = c$ ; *where c is a constant*
   $\Delta\tau_{ij} = 0$ ; *Actual pheromone accumulator*
3. *For every C cycle, do:*
   3.1 *Random assignment of the first operation*
   3.2 *Define the decidability rule for each ant k*
   3.3 *while* $tabu_k$ *is not full, do:*
   *For every ant k, do:*
   *Determine the set of operations achievable from the current node*
   *Select the next operation to be visitet according to equation* 1
   *Move the ant to the selected operation*
   *Save the selected operation in* $tabu_k$
   3.4 *For every ant k do:*
   *Get* $L_k$ *makespan of the built plan*
   *Save the plan with the lowest makespan of Cycle C*
   *For every* $a_{ij}$ *edge, do:*
   *Calculate* $\Delta\tau_{ij}$ *according to equation* 3 *or* 4
   3.5 *For every* $a_{ij}$ *edge, do:*
   *Update pheromone* $\tau_{ij}$ *according to equation* 2
   $\Delta\tau_{ij} = 0$ ;
   3.6 *Show the shortest plan of C cycle*
   3.7 $tabu_k = \phi$ ; *empty the visited list*
   *End − C Cycle*
4. *Show plan with the shortest makespan*
*End − Procedure*

Pseudocode Description:

1. The best results were obtained with the parameters initialized in $\alpha = 0.2$, $\beta = 0.8$ y $\rho = 0.7$, the number of cycles (or iterations) is fixed at 1000 and the amount of ants (K) is calculated according to the number of jobs, which is the amount elements that the J set has, thereby:

$$K = \frac{|J|}{2}$$

**Equation 5**

2. The pheromone trail of all edges is started in a small positive constant.

3. The Probabilistic Construction Phase of solutions begins by K ants.

3.1 The first operation is selected randomly between nodes initially visited according to the constraints of the problem.

3.2 The selection of the decidability rule is done randomly, with equal probability between the rule with the shortest processing time SPT (Shortest Processing Time) or the rule with the longest processing time LPT (Longest Processing Time) of the operations [30].

3.3 While tabú$_k$ memory has not finished filling, it means that the ant has not completed the plan generation therefore it continuous traveling the graph until completing the total operations

($|O| = |J| * |M|$). The tabu$_k$ list restricts the choice of operations to prevent a return to recently visited nodes. In the set of visited operations are included operations that generate a delay in the machines less or equal to five time units. To maintain the balance affected by the delay generated, visibility of the node is reduced on a percentage point per unit of time lost.

3.4 Once each ant has built a solution, the pheromone actualization process is started, reviewing the traveled path to add the appropriate amount of pheromone according to equation 3 or 4, to the pheromone accumulator of the current cycle. If the makespan of the solution is expensive, less importance is given to the way, thus depositing few pheromone on edges.

3.5 Update pheromone trails of the visited edges using a process known as *posteriori online update*, which is a global update performed offline, that is, after the execution of each cycle of the algorithm. It is deposited in the pheromone trails of each of the edges of the graph, what the ants have been added in the respective pheromone accumulator. Then the actual pheromone accumulator is restarted at zero for not to redeposit this pheromone in the next cycle.

3.6 The best quality plan of the current cycle is saved with its respective makespan.

3.7 Memory (tabu$_k$) is erased on each ant to start building new plans in the next cycle.

4. Shows the best plan of all cycles performed by the algorithm.

## 6. ANALYSIS AND COMPARISON OF RESULTS

Results shown in Table 3 were obtained in 30 executions of the algorithm (1000 iterations) for each of the 40 JSSP instances raised by Lawrence [11], that are of different sizes and difficulty, and because of its wide use, we can compare the results with other techniques that generate the best known solution (BKS) taken from [13] and [27] The table shows, the name of the instance of Lawrence, its size and BKS, the best makespan found and their percentage relative error respect al BKS, the makespan average, standard deviation, and finally the average number of evaluations of the objective function.

Table 3. Experimental results

| Instance | Size | BKS | Best $C_{max}$ | Relative Error (%) | $C_{max}$ Average | Standard deviation | #Eval. Average |
|---|---|---|---|---|---|---|---|
| LA01 | 10 x 5 | 666 | 666 | 0 | 667.8 | 2.3 | 2375 |
| LA02 | 10 x 5 | 655 | 669 | 2.13 | 689.7 | 6.2 | 2809 |
| LA03 | 10 x 5 | 597 | 623 | 4.36 | 644.8 | 8.0 | 2230 |
| LA04 | 10 x 5 | 590 | 611 | 3.56 | 617.7 | 5.0 | 2257 |
| LA05 | 10 x 5 | 593 | 593 | 0 | 593.0 | 0.0 | 101 |
| LA06 | 15 x 5 | 926 | 926 | 0 | 926.0 | 0.0 | 531 |
| LA07 | 15 x 5 | 890 | 890 | 0 | 898.2 | 5.6 | 3443 |
| LA08 | 15 x 5 | 863 | 863 | 0 | 863.1 | 0.4 | 2251 |
| LA09 | 15 x 5 | 951 | 951 | 0 | 951.0 | 0.0 | 391 |
| LA10 | 15 x 5 | 958 | 958 | 0 | 958.0 | 0.0 | 637 |
| LA11 | 20 x 5 | 1222 | 1222 | 0 | 1222.0 | 0.0 | 1504 |
| LA12 | 20 x 5 | 1039 | 1039 | 0 | 1039.0 | 0.0 | 1752 |
| LA13 | 20 x 5 | 1150 | 1150 | 0 | 1150.0 | 0.2 | 2952 |
| LA14 | 20 x 5 | 1292 | 1292 | 0 | 1292.0 | 0.0 | 471 |
| LA15 | 20 x 5 | 1207 | 1212 | 0.41 | 1245.6 | 9.9 | 3836 |
| LA16 | 10 x 10 | 945 | 1005 | 6.35 | 1020.1 | 10.1 | 2700 |
| LA17 | 10 x 10 | 784 | 812 | 3.57 | 836.1 | 10.9 | 2401 |
| LA18 | 10 x 10 | 848 | 885 | 4.36 | 904.8 | 9.8 | 2946 |

| LA19 | 10 x 10 | 842 | 875 | 3.92 | 881.7 | 4.7 | 2394 |
|------|---------|-----|-----|------|-------|-----|------|
| LA20 | 10 x 10 | 902 | 912 | 1.11 | 936.8 | 9.6 | 2496 |
| LA21 | 15 x 10 | 1046 | 1107 | 5.38 | 1162.3 | 16.4 | 3658 |
| LA22 | 15 x 10 | 927 | 1018 | 9.82 | 1050.2 | 14.2 | 2938 |
| LA23 | 15 x 10 | 1032 | 1051 | 1.84 | 1069.2 | 10.0 | 3826 |
| LA24 | 15 x 10 | 935 | 1011 | 8.13 | 1033.5 | 8.3 | 3097 |
| LA25 | 15 x 10 | 977 | 1062 | 8.7 | 1093.3 | 14.1 | 3632 |
| LA26 | 20 x 10 | 1218 | 1296 | 6.4 | 1339.6 | 16.2 | 5955 |
| LA27 | 20 x 10 | 1235 | 1362 | 10.28 | 1379.8 | 9.6 | 4450 |
| LA28 | 20 x 10 | 1216 | 1330 | 9.38 | 1363.8 | 13.9 | 3938 |
| LA29 | 20 x 10 | 1157 | 1339 | 15.73 | 1374.4 | 11.9 | 4532 |
| LA30 | 20 x 10 | 1355 | 1410 | 4.06 | 1443.2 | 15.0 | 5186 |
| LA31 | 30 x 10 | 1784 | 1798 | 0.78 | 1825.8 | 12.5 | 7098 |
| LA32 | 30 x 10 | 1850 | 1868 | 0.97 | 1906.0 | 20.7 | 8016 |
| LA33 | 30 x 10 | 1719 | 1731 | 0.7 | 1771.0 | 15.1 | 5796 |
| LA34 | 30 x 10 | 1721 | 1788 | 3.89 | 1823.9 | 13.9 | 6811 |
| LA35 | 30 x 10 | 1888 | 1913 | 1.32 | 1974.1 | 22.8 | 7357 |
| LA36 | 15 x 15 | 1268 | 1396 | 10.09 | 1430.4 | 18.5 | 3405 |
| LA37 | 15 x 15 | 1397 | 1517 | 8.59 | 1544.2 | 12.7 | 2142 |
| LA38 | 15 x 15 | 1196 | 1315 | 9.95 | 1343.8 | 10.1 | 4051 |
| LA39 | 15 x 15 | 1233 | 1304 | 5.76 | 1359.5 | 16.2 | 3266 |
| LA40 | 15 x 15 | 1222 | 1307 | 6.96 | 1323.7 | 9.2 | 2655 |
| | | | Average: | 3.96 | | 9.09 | 3307.15 |

Although the efficacy of the algorithm to find the optimum isn't high, reaching the BKS in 27.5% of the LA instances, the average relative error in the 40 instances is only 4%, which is a good approximation to the optimal of JSSP. And the highlight is the low average number of objective function evaluations, which is much lower compared to the methods that obtained the BKS (Table 4). The AIS on average takes 52 times more evaluations our algorithm and Cultural algorithm (CULT) has 137 times more evaluations. Compared with Tabu Search (TS) the number of evaluations is only three times lower, because it is not included the number of evaluations performed by the INSA algorithm that gives the TS base solution [28]. So we can state that our algorithm has a high computational efficiency, reducing costs in time and memory, something very important in this type of problems where getting an "economic" solution is as important as the quality of it. Furthermore, EAS is an algorithm stable because its standard deviation is very low.

The following table compares the average number of objective function evaluations made by EAS, with those made by TS, AIS and CULT [29]:

Table 4. Number of evaluations of objective function

| Algorithm | N° of Average evaluations |
|-----------|---------------------------|
| EAS | 3307 |
| AIS | 175058 |
| CULT | 454525 |
| TS | 11108 |

Table 3 shows that on the problems of size 10 x 5 did not have trouble finding the BKS, with the exception of the instance LA04 until LA02l, where the best obtained result is close to the

BKS (less than 5%). Also for instances of size 15 x 5 and 20 x 5, with the exception of the LA15 that only moves away from BKS in 5 units of time. In the other instances (size 10 x 10, 15 x 10, 20 x 10, 30 x 10 and 15 x 15), which have 5 or 10 machines more than the previous, complexity is quite high because of the considerable number of operations to be performed, this means lower quality solutions obtained. For example, to instances of size 30 x 10, 300 operations must be performed, and the total number of possible combinations is $(30!)^{10}$, that is approximately $2.65 \times 10^{42}$. However, the algorithm achieves to present high quality solutions on instances of 30 x 10 (Figure 5). In general, 65% of executed instances approaches less than 5% of BKS and 47.5% deviate by less than 3% of the BKS.

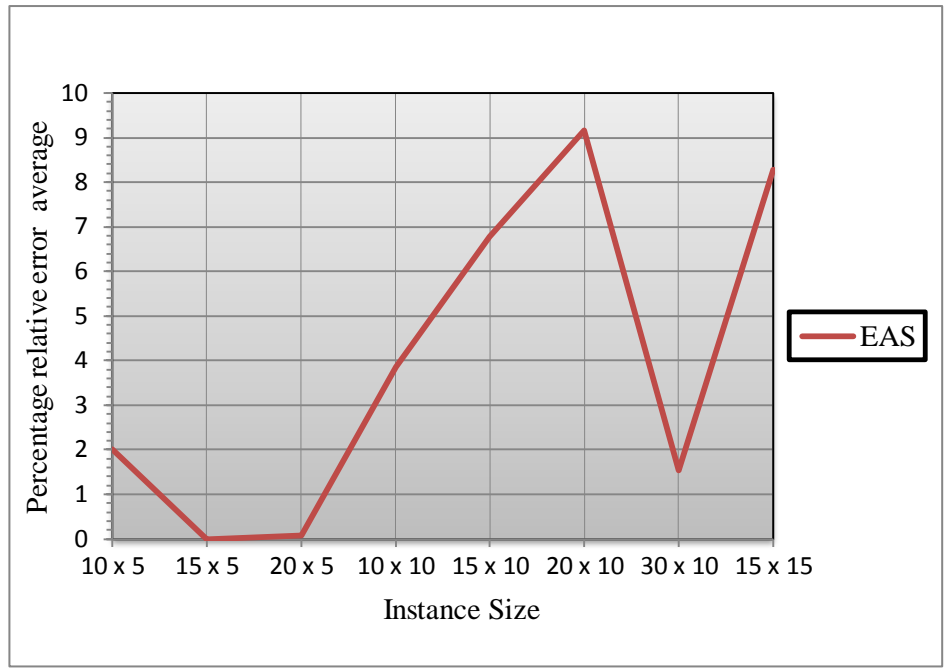

Figure 5. Relative Error average by instance size

## 7. CONCLUSIONS

The Ant Colony Optimization is a technique of swarm intelligence, which is applied for combinatorial optimization problems as JSSP. The algorithm implemented, Elitist Ant System, has proven to be competitive by find good quality solutions to JSSP in a low number of objective function evaluations, although requires improvements to obtain the best known solution in all LA instances. Therefore, ACO is a metaheuristic that has the potential to obtain efficiently solutions of scheduling problems, with minimal cost of time and computational resources.

**Authors**

Edson Flórez was born at San Gil (Colombia) in 1991, is currently an engineer student at School of Engineering and Computing Systems of the Universidad Industrial de Santander. His research areas are on Swarm Intelligence algorithms, Operations Research and Distributed systems.

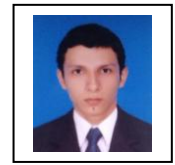

Wilfredo Gómez was born at Bucaramanga(Colombia) in 1984, is currently an Magister candidate in system engineering at School of Engineering and Computing Systems of the Universidad Industrial de Santander. His research areas are on Bioinspired Computation, Operations Research and Education.

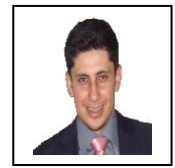

Lola Bautista is MSc. in Computer Engineering University of Puerto Rico she is currently the Director of the Research Group in Biomedical Engineering. His research areas are on Software Design in Cardiology and Electrocardiography, Digital Signal Processing and Graphics, Data Mining.

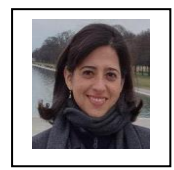